# Enhanced Rooftop Solar Panel Detection by Efficiently Aggregating Local Features

Kuldeep Kurte, Kedar Kulkarni

Reliance Industries Limited, India

{kuldeep.kurte, kedar.kulkarni}@gmail.com

## ABSTRACT

In this paper, we present an enhanced Convolutional Neural Network (CNN)-based rooftop solar photovoltaic (PV) panel detection approach using satellite images. We propose to use pre-trained CNN-based model to extract the local convolutional features of rooftops. These local features are then combined using the Vectors of Locally Aggregated Descriptors (VLAD) technique to obtain rooftop-level global features, which are then used to train traditional Machine Learning (ML) models to identify rooftop images that do and do not contain PV panels. On the dataset used in this study, the proposed approach achieved rooftop-PV classification scores exceeding the predefined threshold of 0.9 across all three cities for each of the feature extractor networks evaluated. Moreover, we propose a 3-phase approach to enable efficient utilization of the previously trained models on a new city or region with limited labelled data. We illustrate the effectiveness of this 3-phase approach for multi-city rooftop-PV detection task.

## CCS CONCEPTS

• Computing methodologies → Machine learning

## KEYWORDS

Rooftop-PV detection, Convolutional Neural Network, VLAD, ML

## 1 Introduction

Accurate information on existing solar photovoltaic (PV) panels on building rooftops is essential for effective solar capacity planning, allowing manufacturers to target marketing efforts in areas with many no-rooftop-PV buildings. While manual data collection can provide such information, it is often time-consuming, labor-intensive, and costly [1]. In contrast, utilizing satellite or aerial imagery for rooftop-PV detection offers a scalable and cost-effective solution. This paper focuses on detecting buildings with and without rooftop-PVs using high-resolution satellite images.

Most rooftop-PV detection studies in the literature focus on segmenting rooftop-PV panels using high-resolution satellite imagery and hand-labeled training data of rooftop-PV, which are costly and labor-intensive to produce [2-6]. As a result, these studies were often limited to specific regions or cities, hindering performance assessment for multi-city rooftop-PV detection tasks. An exception is DeepSolar++ [2], which trained a CNN-based rooftop-PV detection model using approximately 18,000 high-resolution satellite images from 50 U.S. cities. However, such extensive datasets may not be readily available in other countries with diverse rooftops and PV patterns, making it difficult to train CNN-based models with limited labeled data.

We propose an approach to efficiently utilize the limited labelled dataset of rooftops (with and without PV) for a multi-city rooftop-PV detection task. Typically, PV panels occupy a specific portion of the rooftop area, appearing homogeneous in texture and color in the satellite image, when compared to the portion that does not have PV panels [1]. To exploit this image-level homogeneity of rooftop-PVs, we use pre-trained CNN models to extract local features. Next, we employ VLAD [7] to fuse local features into fixed-length global features per rooftop, which are then used to train traditional ML models for rooftop-PV classification. Further, the trained rooftop-PV classification models can be transferred across cities in a stepwise manner, based on the proposed 3-phase approach, by efficiently using new (limited) labelled data and previously trained models. More details on the 3-phase approach are provided in Section II.

Our work differs from previous approaches in the following ways: 1) we formulate the problem as a rooftop image classification task as opposed to a segmentation task, 2) we assume limited availability of labelled data, which, in practice, is typically obtained incrementally, i.e., one city at a time; 3) we propose VLAD-based fusion of sub-rooftop-level local features; and 4) we propose a 3-phase approach to efficiently utilize a pre-trained rooftop-PV model on multiple cities with limited labels.

This paper is organized as follows: Section 2 outlines two rooftop-PV classification approaches and introduces a 3-phase method for extending rooftop-PV classification to multiple cities. Section 3 presents benchmarking results for these methods, while Section IV offers concluding remarks and future directions for rooftop-PV detection.

## 2 Methodology

In this work, we evaluated two approaches for rooftop-PV classification, as shown in Fig. 1.

**Building-Rooftop (BR-ML) classification**: In this approach, a building's rooftop image is classified as either "with-PV" or "with-no-PV". The rooftop images were obtained by clipping high-resolution satellite images using the corresponding building footprint geometries. As a result, the rooftop images vary in size. The BR-ML approach first resize variable-sized images to a common size and then use a pre-trained CNN model as a feature extractor to extract the features from each of the rooftop images. Further, the extracted features were used to train ML models to classify rooftop images.

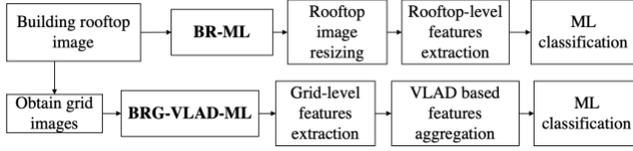

**Figure 1: Approaches used for rooftop PV detection**

**Building-Rooftop-Grid (BRG-VLAD-ML) classification**: Resizing rooftop images in BR-ML may alter the appearance of a rooftop in terms of its size and shape. This image-level distortion can result in the loss of local information related to the size and shape of PV panels. The BRG-VLAD-ML approach aims to obtain global (rooftop-level) features by efficiently fusing localized features extracted from smaller portions of the rooftop images. This approach has three steps.

*Step 1 - Grid-level feature extraction*: In this step, a building rooftop image is divided into a set of regular-sized, non-overlapping grid images. A grid image that largely covers a PV panel appears more homogeneous in color and texture, showing a strong signature of the presence of a PV panel. In contrast, a grid image with the partial PV coverage has presence of both PV as well as non-PV areas. Both the types of grid images, i.e., with the prominent PV presence and with the partial PV coverage were considered as grid images with-PV. Similar to the BR-ML, we use a pre-trained CNN model to extract grid-level features.

*Step 2 - Vectors of Locally Aggregated Descriptors (VLAD)-based aggregation:* The BRG-VLAD-ML approach utilizes the VLAD technique to combine grid-level features into rooftop-level descriptors. Given that the number of grid images from each rooftop can vary, concatenating grid-level features would lead to variable-length descriptors, which are unsuitable for training ML models for rooftop-PV classification. VLAD efficiently fuses local features from various grid images into fixed-length global feature descriptors, ensuring uniformity across all rooftop images [7].

*Step 3 - ML classification*: We used the rooftop-level VLAD features generated in step 2 to train ML models such as Logistic Regression (LR) classifier, Random Forest (RF) classifier, and Support Vector Classifier (SVC) for rooftop classification. Since these ML models are computationally less expensive to train, compared to fine tuning the CNN models, we expect a significant gain in terms of execution times of the BRG-VLAD-ML based rooftop classification models.

The rooftop-PV classification models of the categories BR-ML or BRG-VLAD-ML that are trained and validated on one city can be adapted (fine-tuned or re-trained) for any new city for rooftop-PV classification. To make a systematic use of pre-trained rooftop-PV detection models for any new city with limited labelled data, we proposed a 3-phase approach, described as follows:

**Phase-1 (Evaluation)**: In this phase, we evaluate the performance of the previously trained rooftop-PV classification models using the combined test data of the previous as well as new city. If the accuracy of the previous pre-trained model is above the acceptable threshold, then we stop the process and record the execution times. Further, we use the pre-trained model to classify all the rooftop images from the new city.

**Phase-2 (Tuning Hyperparameters)**: If the Phase-1 accuracy of the best pre-trained classification model is below the acceptable threshold, we move to Phase-2. In this phase, we validate the performance of the pre-trained rooftop-PV classification models for all combinations of the hyperparameters, which were stored in the Phase-3 of a previous city. We used a combined training data from past cities and the new city as a validation-set to search the best Phase-2 model. The final accuracy of the best Phase-2 model is computed using the remaining combined test data of the all the cities, including the new city. If the test accuracy of the best performing Phase-2 model is above the acceptable threshold, then we stop the process otherwise we go to Phase-3 for active training.

**Phase-3 (Active training):** In this phase, we combine training data from all cities, including new city to train new ML models for rooftop-PV classification. Further the combined test data was used as to evaluate the final performance of the newly trained model in Phase-3. In both BR-ML and BRG-VLAD-ML approaches, we train a new set of Phase-3 ML models for each new city. We perform HPO to tune hyperparameters specific to the ML algorithm. In BRG-VLAD-ML approach, we additionally tune grid size as well as VLAD parameter 'K', i.e., number of clusters.

During all three phases whenever the accuracy of rooftop-PV classification surpasses a pre-defined threshold we stop the process and record the total execution time.

## 3  Results and Discussions
## 3.1  Dataset used

To evaluate multi-city rooftop-PV detection, we obtained high-resolution rooftop images via the Google Static Map API [9] for three regions in the state of Maharashtra, India: Reliance Corporate Park (RCP) campus, Chakan town, and Pune city. Using internal building footprint data, we manually labelled a few buildings from each city/region as "with-PV" or "with-no-PV." Table 1 shows class distributions by the number of rooftops and grid images (in brackets) for each city. We applied data augmentation using the "imgaug" Python library [10], incorporating techniques such as horizontal and vertical flips, random cropping, gamma contrast adjustment, Gaussian blurring, brightness modification, and affine transformations (rotation and shearing) to balance the number of rooftops labeled as "with-PV" (minority class) with those labeled as "no-PV".

**Table 1: Distributions of building rooftops and grid images "with-PV" and "with-no-PV" across the three regions**

| City | "with-PV" rooftops (grids) | "with-no-PV" rooftops (grids) | Total rooftops (grids) |
|---|---|---|---|
| **RCP** | 42 (394) | 507 (3356) | 549 (3750) |
| **Chakan** | 73 (2880) | 98 (14212) | 171 (17092) |
| **Pune** | 195 (8913) | 195 (8247) | 390 (17160) |

## 3.2 Experimental Setup

All experiments were conducted on an Apple MacBook Pro M1 CPU, with 16GB of memory, and 500GB of storage. We used pre-trained models such as MobileNetV2 [11], VGG16 [11], Densenet121 [11], and DeepSolar++ [2] as backbone networks for feature extraction for training both BR and BRG based approaches. We used libraries such as Keras-Tuner [12] for BR model training and HPO, rasterio [13] and GeoPandas [14] for geospatial processing, the VLAD library [15] for VLAD training, and scikit-learn [16] for training and HPO of the ML models.

Table 2 presents the data used for training, validation, and testing throughout all three phases across all cities. For the first region, i.e., RCP, we proceeded directly to Phase-3 where we trained ML models for different sets of hyperparameters. Table 3 presents the total number of ML models trained, based on the specific hyperparameters and their corresponding values for each model. For instance, for LR model, we evaluated three values for C and two for the *solver*, yielding six hyperparameter combinations and resulting in six trained ML models for the BR-ML approach. In BRG-VLAD-ML, varying the *grid size* (3 values), VLAD parameter *K* (3 values), and the 9 C and *solver* combinations produced a total of 54 hyperparameter combinations.

**Table 2: Data used for training, validation, and testing**

| City | Phase-1 (Evaluation) | Phase-2 (Tuning Hyperparameters) | Phase-3 (Active training) |
|---|---|---|---|
| RCP | - | - | Train+HPO: RCP train<br>Test: RCP test |
| Chakan | Test: Test set of RCP+Chakan | HPO evaluation: Train set of RCP+Chakan<br>Test: Test set of RCP+ Chakan | Train+HPO: Train set of RCP+ Chakan<br>Test: Test set of RCP+ Chakan |
| Pune | Test: Test set of RCP+Chakan + Pune | HPO evaluation: Train set of RCP+Chakan+Pune<br>Test: Test set of RCP+ Chakan+Pune | Train+HPO: Train set of RCP+ Chakan+Pune<br>Test: Test set of RCP+Chakan+Pune |

**Table 3: Hyperparameter values used for ML models**

| Classification models | Hyperparameters | Number of Models trained in Phase-3 |
|---|---|---|
|  | Grid size [64, 96, 128] | - |
|  | K [2, 3, 4] | - |
| LR | C: [0.01, 0.1, 1, 10],<br>*solver*: ['liblinear', 'lbfgs'] | BR-ML: 6<br>BRG-VLAD-ML: 54 |
| RF | *n_estimators*: [50, 100, 200],<br>*max_depth*: [None, 10, 20] | BR-ML: 9<br>BRG-VLAD-ML: 81 |
| SVC | C: [0.1, 1, 10 ]<br>*kernel*: ["linear", "rbf"] | BR-ML:6<br>BRG-VLAD-ML: 54 |

We calculated F1-scores at both city-level as well as at a global-level (i.e., considering all cities together). For example, for Chakan, we computed city-level F1- scores using RCP-test set and Chakan-test set separately and computed global F1-score using the combined test sets of RCP and Chakan. We further obtained a weighted F1-score by taking a weighted summation of the average F1-scores at city-level and the global F1-score. This weighted F1-score reflects the model's performance at the city-level while simultaneously accounting for its ability to generalize across all regions. In this work, we used equal weights of 0.5 while summing both global as well as average of the city-level F1-scores. Based on our internal business requirement, we used 0.90 (up to 2 decimal) as a pre-defined threshold for the combined F1-score as a stopping criterion across 3 phases, including 3 stages of active training of Phase-3. Along with the F1-scores, we measured the total time taken by an approach to surpass the 0.90 threshold at any given phase for a specific city.

## 3.3 Results and discussion

Table 4 summarizes the weighted F1-scores and execution times for the BR-ML and BRG-VLAD-ML methods used in rooftop-PV classification. These two approaches were assessed using four different pre-trained backbone networks for feature extraction. The shaded cells indicate cases where the weighted F1-scores fell below the specified threshold. The BRG-VLAD-ML method consistently achieved weighted F1-scores above 0.9 across all three cities for all four backbone networks. However, the total execution time for BRG-VLAD-ML was longer than for BR-ML. This increase in execution time can be attributed to: 1) the longer feature extraction time due to the larger number of grid images compared to rooftop images (as shown in Table 1), and 2) the additional time required to fine-tune the grid size and optimize "K", the VLAD parameter.

**Table 4: Weighted F1-scores and execution times of BR-ML and BRG-VLAD-ML approaches across different cities with different backbone architectures**

|  | Backbone | RCP | RCP + Chakan | RCP + Chakan + Pune | Total time (mins.) |
|---|---|---|---|---|---|
| BR-ML | MobileNetV2 | 0.96 | 0.92 | 0.86 | 2.5 |
|  | Densenet121 | 0.96 | 0.97 | 0.91 | 2.9 |
|  | VGG16 | 0.83 | 0.86 | 0.84 | 5.3 |
|  | DeepSolar++ | 1.00 | 0.92 | 0.88 | 5.3 |
| BRG-VLAD-ML | MobileNetV2 | 1.00 | 0.97 | 0.92 | 15.8 |
|  | Densenet121 | 1.00 | 0.93 | 0.96 | 35.5 |
|  | VGG16 | 1.00 | 0.97 | 0.95 | 56.9 |
|  | DeepSolar++ | 1.00 | 0.95 | 0.92 | 96.9 |

For conciseness, we chose BR-ML with Densenet121 and BRG-VLAD-ML with MobileNetV2 from Table 4 to demonstrate their city-level and global F1-scores. The Densenet121 was selected for BR-ML due to its weighted F1-score exceeding 0.9 across all three cities. For BRG-VLAD-ML, since all four models achieved weighted F1-scores above 0.9, we selected MobileNetV2. Table 5 highlights that the BRG-VLAD-ML model outperformed BR-ML by achieving a higher global F1-score, while also maintaining city-level F1-scores above 0.90, specifically for RCP and Pune. We compared the city-level and global F1-scores of the BR-ML and BRG-VLAD-ML methods across all four backbone networks. Overall, the BRG-VLAD-ML approach consistently delivered superior performance compared to BR-ML, both in city-specific and global F1-scores across all cities and backbones. In terms of execution time, MobileNetV2 demonstrated superior performance among the four architectures due to its reduced number of parameters relative to the others [11].

**Table 5: Comparison of city-specific F1-scores of BR-ML and BRG-VLAD-ML models**

| Current City | Current + previous cities | BR-ML (Densenet121) | | BRG-VLAD-ML (MobileNetV2) | |
|---|---|---|---|---|---|
| | | City | Global | City | Global |
| RCP | RCP | 0.96 | 0.96 | 1.0 | 1.0 |
| Chakan | RCP | 1.00 | 0.96 | 1.0 | 0.96 |
| | Chakan | 0.94 | | 0.94 | |
| Pune | RCP | 1.0 | 0.89 | 0.91 | 0.92 |
| | Chakan | 0.88 | | 0.98 | |
| | Pune | 0.88 | | 0.89 | |
| Total time (min.) | | 2.9 mins. | | 15.8 mins. | |

We extended our analysis of the BR-ML and BRG-VLAD-ML approaches by comparing them with other grid-level feature aggregation methods, including Fisher vector (BRG-FV-ML) [17] and feature averaging (BRG-AVG-ML) that averages the grid-level bottleneck features. The Table 6 shows the weighted F1-scores of these two models with all four backbones. For both the approaches Densenet121 showed best weighted F1-scores. However, we observed several instances (in gray) where the weighted F1-scores dropped below the predefined threshold, in contrast to the weighted F1-scores of BRG-VLAD-ML which is always above 0.9 across all cities and for all four backbones, as shown in Table 5. Overall, we observed that for grid-level feature aggregation, the VLAD approach consistently outperformed the other methods considered in this study for rooftop-PV classification.

**Table 6: Weighted F1-scores and execution times of BRG-FV-ML and BRG-AVG-ML with four backbone networks.**

| | Backbone | RCP | RCP + Chakan | RCP + Chakan + Pune | Total time (mins.) |
|---|---|---|---|---|---|
| BRG-FV-ML | MobileNetV2 | 0.96 | 0.88 | 0.90 | 19.4 |
| | Densenet121 | 0.96 | 0.95 | 0.92 | 34.7 |
| | VGG16 | 0.92 | 0.88 | 0.89 | 58 |
| | DeepSolar++ | 1.00 | 0.92 | 0.94 | 107.8 |
| BRG-AVG-ML | MobileNetV2 | 0.88 | 0.96 | 0.90 | 12.4 |
| | Densenet121 | 1.00 | 0.97 | 0.91 | 31.6 |
| | VGG16 | 0.81 | 0.95 | 0.91 | 54.1 |
| | DeepSolar++ | 0.89 | 0.93 | 0.93 | 96.3 |

We further benchmarked the proposed approach against recent architectures such as Vision Transformer (ViT) [18], ConvNextV2 [19], CLIP + ViT for zero-shot learning [20], and SegFormer [21]. The active training in Phase-3 for ViT and ConvNextV2 was done in three stages: in stage-1, the final classification layer was re-trained; in stage-2, the preceding layer was re-trained; and in stage-3, a 128-unit dense layer was added and trained alongside the final classification layer. The Phase-3 was stopped at any stage when the accuracy surpassed the 0.9 threshold. The training data used for BR-ML and BRG-VLAD-ML approaches, as shown in Table 2, was also used for ViT and ConvNextV2. For SegFormer, the classification layer was unfrozen and re-trained in stage-1, the linear fusion layer in stage-2, and the entire decode head in stage-3. A heuristic was employed to classify rooftop PV presence by calculating the percentage of PV-marked pixels in the binary mask output and comparing it to a threshold (10%-20%), tuned with validation data. For SegFormer training, around 20 images (10 with PV, 10 without) were randomly selected from each city to generate masks, with 6 rooftop images (3 with PV, 3 without) used as a validation set from each city. The final evaluation was performed using the test dataset shown in Table 2. Due to the computational resource constraints, we trained SegFormer model on a system with Intel Xeon 6240R CPU with 64GB of RAM instead of on MacBook system described in Sec. 3.2. From Table 7, we can observe that these advanced models take significantly high training time as compared with BR and BRG based approaches. For the data used in this study, the combined scores of BR and BRG based techniques are better than the combined scores of these advanced models for rooftop-PV classification task.

**Table 7: Weighted F1-scores and execution times of the advanced approaches for rooftop-PV classification**

| City | RCP | RCP + Chakan | RCP + Chakan + Pune | Total Time (mins.) |
|---|---|---|---|---|
| ViT | 0.89 | 0.95 | 0.93 | 526 |
| ConvNextV2 | 0.96 | 0.90 | 0.89 | 38.8 |
| CLIP(zero-shot) | 0.41 | 0.50 | 0.60 | 47 |
| SegFormer | 1.00 | 0.90 | 0.85 | 121 |

To evaluate the generalization of the trained rooftop-PV models, we tested the best models across all the model categories that are considered in this work using 50 rooftop samples (25 with PV and 25 without PV) from Kochi, Kerala, a city in southern India. Table 8 presents the corresponding F1 scores, where BRG-VLAD-ML and BRG-FV-ML demonstrated the highest performance. To further validate the generalizability of the proposed approach, we plan to extend the evaluation to cities in the eastern and northern regions of India.

**Table 8: Evaluation scores of various models on Kochi data**

| F1 score | BR-ML | BRG-VLAD-ML | BRG-FV-ML | BRG-AVG-ML | ConvNextV2 | ViT | SegFormer |
|---|---|---|---|---|---|---|---|
| Kochi | 0.91 | 0.94 | 0.96 | 0.89 | 0.92 | 0.9 | 0.86 |

## 4 Conclusions

In this work, we demonstrated BRG-VLAD-ML approach for rooftop-PV classification, which decomposes rooftop image into regular-sized grid images. It extracts the grid-level local features using pre-trained CNN models, fuses them using VLAD technique to obtain rooftop-level global features, and further train ML models for rooftop classification. This approach achieved a better weighted F1-score compared to the BR-ML approach while maintaining its city-level F1-scores well above the threshold of 0.9 across all the cities for each of the backbone networks considered in this work. Additionally, we proposed a 3-phase method to efficiently use pre-trained models for multi-city rooftop-PV detection with limited labelled data. In the future, we plan to evaluate the generalization capability of BRG-VLAD-ML approach across a larger number of cities.